\documentclass[10pt,journal,compsoc]{IEEEtran}

\usepackage[ruled]{algorithm2e}

\usepackage{framed}
\usepackage[ruled]{algorithm2e}
\usepackage{amssymb}
\usepackage{latexsym}
\usepackage{epsfig}
\usepackage{graphicx}
\usepackage{booktabs,multirow, multicol}
\usepackage{epsfig}
\usepackage{caption}
 
\usepackage{epstopdf}
\usepackage{hyperref}

\ifCLASSOPTIONcompsoc

  \usepackage[nocompress]{cite}
\else
  \usepackage{cite}
\fi

\ifCLASSINFOpdf

\else

\fi

\begin{document}

\title{A Score-level Fusion Method for Eye Movement Biometrics}

\author{Anjith~George,~\IEEEmembership{Member,~IEEE,}
        and~Aurobinda~Routray,~\IEEEmembership{Member,~IEEE}
\IEEEcompsocitemizethanks{\IEEEcompsocthanksitem A. George and A. Routray are with the Department
of Electrical Engineering, IIT Kharagpur, India,
721302.\protect\\

E-mail: anjith2006@gmail.com
\IEEEcompsocthanksitem This is the author version of the accepted manuscript. The full version of the method described in this paper is available in : Anjith George and Aurobinda Routray. "A Score-level Fusion Method for Eye Movement Biometrics, Pattern Recognition Letters. http://www.sciencedirect.com/science/article/pii/S0167865515004067}
\thanks{DOI: http://dx.doi.org/10.1016/j.patrec.2015.11.020}}

\markboth{}%
{}

\IEEEcompsoctitleabstractindextext{%
\begin{abstract}
This paper proposes a novel framework for the use of eye movement patterns for biometric applications. Eye movements contain abundant information about cognitive brain functions, neural pathways, etc. In the proposed method, eye movement data is classified into fixations and saccades. Features extracted from fixations and saccades are used by a Gaussian Radial Basis Function Network (GRBFN) based method for biometric authentication. A score fusion approach is adopted to classify the data in the output layer. In the evaluation stage, the algorithm has been tested using two
types of stimuli: random dot following on a screen and text reading. The results indicate the strength of eye movement pattern as a biometric modality.  The algorithm has been evaluated on BioEye 2015 database and found to outperform all the other methods. Eye movements are generated by a complex oculomotor plant which is very hard to spoof by mechanical replicas. Use of eye movement dynamics along with iris recognition technology may lead to a robust counterfeit-resistant person identification system.
\end{abstract}

\begin{IEEEkeywords}
Eye tracking, Biometrics, Eye movement biometrics, Gaze tracking.
\end{IEEEkeywords}}

\maketitle

\IEEEdisplaynontitleabstractindextext

\IEEEpeerreviewmaketitle

\IEEEraisesectionheading{\section{Introduction}\label{sec:introduction}}

\IEEEPARstart{B}{iometrics} is an active area of research in pattern recognition and machine learning community. Potential applications of biometrics include forensics, law enforcement, surveillance, personalized interaction, access control ~\cite{jain2007handbook}, etc. Physiological features like fingerprint, DNA, earlobe geometry, iris pattern, facial recognition, ~\cite{jain2004introduction} are widely used in biometrics. Recently, several behavioral biometric modalities have been proposed including gait, eye movement patterns, keystroke dynamics ~\cite{wang2009behavioral} signature, etc. Even though many such parameters like brain signals ~\cite{marcel2007person} (using electroencephalogram) and heart beats ~\cite{plataniotis2006ecg} have been proposed as biometric modalities, their invasive nature limits their practical applications.

An effective biometric should have the following characteristics ~\cite{jain2007handbook}: 1) the features should be unique for each individual, 2) they should not change with time (template aging effects), 3) acquisition of parameters should be easy (low computational complexity and noninvasive), 4) accurate and automated algorithms should be available for classification, 5) counterfeit resistance, 6) low cost, and 7) ease of implementation. Other characteristics that might make the system more robust are portability and the ability to extract features from non-co-operative subjects.

Out of many biometric modalities, iris recognition has shown the most promising results ~\cite{ib2005independent} obtaining Equal Error Rates (EER) close to 0.0011\%. However, it can only be used when the user is co-operative. Such systems can be spoofed by contact lenses with printed patterns. Even though most of the biometric modalities perform well on evaluation databases, one may be able to spoof such systems with mechanical replicas or artificially fabricated models ~\cite{roberts2007biometric}. In this regard, several approaches have been presented ~\cite{schuckers2002issues} to detect the liveliness of tissues or body parts presented to the biometric system. However, such methods are also vulnerable to spoofing.

Biometrics using patterns obtained from eye movements is a relatively new field of research. Most of the conventional biometrics use physiological characteristics of the human body. Eye movement-based biometrics tries to identify the behavioral patterns as well as information regarding physiological properties of tissues and muscles generating eye movements ~\cite{leigh1999neurology}. They provide abundant information about cognitive brain functions and neural signals controlling eye movements. Saccadic eye movement is   the fastest movement (peak angular velocities up to 900 degrees per second) in the human body. Mechanically replicating such a complex oculomotor plant model is extremely difficult. These properties make eye movement patterns a suitable candidate for biometric applications. The dynamics of eye movement along with these properties can give inbuilt liveliness detection capability.

Initially, eye movement biometrics has been proposed as a soft biometric. However, with the high level of accuracy achieved, it seems there are more opportunities regarding its application as an independent biometric modality. Eye movement detection can be integrated easily into already existing iris recognition systems. A combination of iris recognition and eye movement pattern recognition may lead to a robust counterfeit-resistant biometric modality with embedded liveliness detection and continuous authentication properties. Eye movement biometrics can also be made task-independent ~\cite{kinnunen2010towards} so that the movements can be captured even for non-co-operative subjects.

The rest of the paper is organized as follows. Section 2 describes previous works related to the use of eye movement as a biometric. Section 3 presents the proposed algorithm. Evaluation of the algorithm along with the results are outlined in section 4. Conclusions regarding eye movement biometrics and possible extensions are detailed in section 5.
\section{Related works}

Initial attempts to use eye movements as a biometric modality were carried out by Kasprowski and Ober ~\cite{kasprowski2004eye}. They recorded the eye movements of subjects following a jumping dot on a screen. Several frequency domain and Cepstral features were extracted from this data. They applied different classification methods like naive Bayes, C45 decision trees, SVM and KNN methods. The results obtained further motivated research in eye movement-based biometrics. Bednarik et al. ~\cite{bednarik2005eye} conducted experiments on several  tasks including text reading, moving cross stimulus tracking and free viewing of images. They used FFT and PCA on the eye movement data. Several combinations of such features were tried. However, the best results were obtained using the distance between eyes, which is not related to eye dynamics. Komogortsev et al. ~\cite{komogortsev2010biometric} used an Oculomotor Plant Mathematical Model (OPMM) to model the complex dynamics of the oculomotor plant. The plant parameters were identified from the eye movement data. This approach was further extended in ~\cite{komogortsev2012biometric}. Holland and Komogortsev ~\cite{holland2013complexb} evaluated the applicability of eye movement biometrics with different spatial and temporal accuracies and various types of stimuli. Several parameters of eye movements were extracted from fixations and saccades. Weighted components were used to compare different samples for biometric identification. A temporal resolution of 250 Hz and spatial accuracy of 0.5 degrees were identified as the minimum requirements for accurate gaze-based biometric systems. Kinnunen et al. ~\cite{kinnunen2010towards} presented a task-independent user authentication system based on eye movements. Gaussian mixture modeling of short-term gaze data was used in their approach. Even though the accuracy rates were fairly low, the study opened up possibilities for the development of task-independent eye movement-based verification systems. Rigas et al. ~\cite{rigas2012biometric} explored variations in individual gaze patterns while observing human face images. Eye movements resulted were analyzed using a graph-based approach. The Multivariate Wald-Wolfowitz runs test was used to classify the eye movement data. This method achieved 70\% rank-1 IR and 30\% EER on a database of 15 subjects. Rigas et al. ~\cite{rigas2012human} extended this method using features of velocity and acceleration calculated from fixations.  The feature distributions were compared using Wald-Wolfowitz test.

Zhang et al. ~\cite{zhang2012biometric} used saccadic eye movements with machine learning algorithms for biometric verification. They used multilayer perceptron networks, support vector machines, radial basis function networks and logistic discriminant for the classification of eye movement data. Recently Cantoni et al. ~\cite{cantoni2015gant} proposed a gaze analysis technique called GANT in which fixation patterns were denoted by a graph-based representation. For each user, a fixation model was constructed using the duration and number of visits at various points. Frobenius norm of the density maps was used to find the similarity between two recordings. Holland and Komogortsev presented an approach (CEM) ~\cite{holland2011biometric} using several scan path features including saccade amplitudes, average saccade velocities, average saccade peak velocities, velocity waveform, fixation counts, average duration of fixation, length of scan path, area of scan path, regions of interest, number of inflections, main sequence relationship, pairwise distances between fixations, amplitude duration relationship, etc. A comparison metric of the features was computed using Gaussian cumulative density function. Another similarity metric was obtained by comparing the scan paths. A weighted fusion of these parameters obtained the best case EER of 27\%. Holland and Komogortsev proposed a method (CEM-B)~\cite{holland2013complexa}, in which the fixation and saccade features were compared using statistical methods like Ansari-Bradley test, two-sample t-test, two-sample Kolmogorov-Smirnov test, and the two-sample Cramer-von Mises test. Their approach achieved 83\% rank-1 IR and 16.5\% EER on a dataset of 32 subjects.

To the best knowledge of the authors, the best case EER obtained is 16.5\% ~\cite{holland2013complexa}. Most of the works presented in the literature were evaluated on smaller databases. The effect of template aging was not considered in these works. For the application of eye movement as a reliable biometric, the patterns should remain consistent with time. In this paper, we try to improve upon the existing methods. The proposed algorithm can reach an EER up to 2.59\% and rank-1 accuracy of 89.54\% in RAN\_30min dataset of BioEye 2015 database ~\cite{bioeye} containing 153 subjects. Template aging effect has also been studied using data taken after an interval of 1 year. The average EER obtained is 10.96\% with a rank-1 accuracy of 81.08\% with 37 subjects.
\section{Proposed method}
In the proposed approach, eye movement data from the experiment are classified into fixations and saccades, and their statistical features are used to characterize each individual. For each individual, the properties of saccades of same durations have been reported to be similar \cite{collewijn1988binocular}. We use this knowledge and extract the statistical properties of the eye movements for biometric identification. Different stages of the algorithm are described below.

\subsection{Data pre-processing and noise removal}
The data contains visual angles in both $x$ and $y$ directions along with stimulus angles. Information about the validity of samples is also available. Eye movement data has been captured at a sampling frequency of 1000Hz. The data obtained is decimated to 250Hz using an anti-aliasing filter. In the proposed feature extraction method, most of the parameters are computed with reference to the screen coordinate system. Hence, in the pre-processing stage, the data obtained is converted to screen
coordinates based on head distance and geometry of the acquisition system as:
\begin{equation}
{x_{screen}} = \left( {{{d * {w_{pix}}} \over w}} \right)\tan ({\theta _x}) + {{{w_{pix}}} \over 2}
\end{equation}
\begin{equation}
{y_{screen}} = \left( {{{d * {h_{pix}}} \over h}} \right)\tan ({\theta _y}) + {{{h_{pix}}} \over 2}
\end{equation}
where, $d,{\theta _x}$ and ${\theta _y}$ denote distance from the screen and visual angles in $x$ and $y$ direction (in radian) respectively. ${x_{screen}}$ and ${y_{screen}}$ denote the position of gaze on the screen. ${w_{pix}},{h_{pix}}$,$w,h$ denote resolution and physical size of the screen in horizontal and vertical directions respectively.

Raw eye gaze positions may contain noise. Most of the features used in this work are extracted from velocity and acceleration profiles. The presence of noise makes it difficult to estimate the velocity and acceleration parameters using differentiation operation. Eye movement signals contain high-frequency components, especially during saccades. High-frequency components would be more prominent in velocity and acceleration profiles ~\cite{harris1984instrument}. Savitzky-Golay filters are useful for
filtering out the noise when the frequency span of the signal is large ~\cite{krishnan2013selection}. They are reported to be optimal ~\cite{savitzky1964smoothing} for minimizing the least-square error in fitting a polynomial to frames of the noisy data. We use this filter with polynomial order of 6 and frame size of 15 in our approach.

\subsection{Eye movement classification and feature extraction }
\subsubsection{Eye movement classification}
The I-VT (velocity threshold) algorithm ~\cite{holland2012biometric}, ~\cite{salvucci2000identifying} is used to classify the filtered eye movement data into a sequence of fixations and saccades (Algorithm 1). Most of the earlier works specify the velocity threshold for angular velocity. The angular velocity computed from the filtered data is used to classify the eye movements. A velocity of 50 degrees/second is used as the threshold in I-VT algorithm.

\begin{algorithm}[h]
\label{alg:eyeclassification}
\KwData{[Time Gazex Gazey]}
\KwResult{Res}
\textbf{Constants} : VT=Velocity threshold,MDF=Minimum duration for fixation\;
\textbf{States} $=$[FIXATION,SACCADE]\;
fixationStart=1\;
Velocity=\textit{smoothDiff}(data)\;
$ N\leftarrow$ Number of samples of data\;

\For {$index \leftarrow$ 1 \textbf{to} N} {

\eIf{Velocity[index] $<$ VT}{
currentState=FIXATION\;
\If{lastState $\neq$ currentState}{
fixationStart = index\;
}
}{
\If{lastState$ =$ FIXATION}{
duration = data(index,1) - data(fixationStart,1)\;
\If {duration $<$ MDF}{

\For {$i \leftarrow$ fixationStart \textbf{to} index}{
res[i]= SACCADE;
}
}
}
currentState=SACCADE\;
}
lastState=currentState\;
res[index]=currentState\;
}
Res$\leftarrow$res\;
\caption{Fixation and Saccade classification algorithm}
\end{algorithm}
\begin{figure}[h]
\begin{center}
\includegraphics[width=0.9\linewidth]{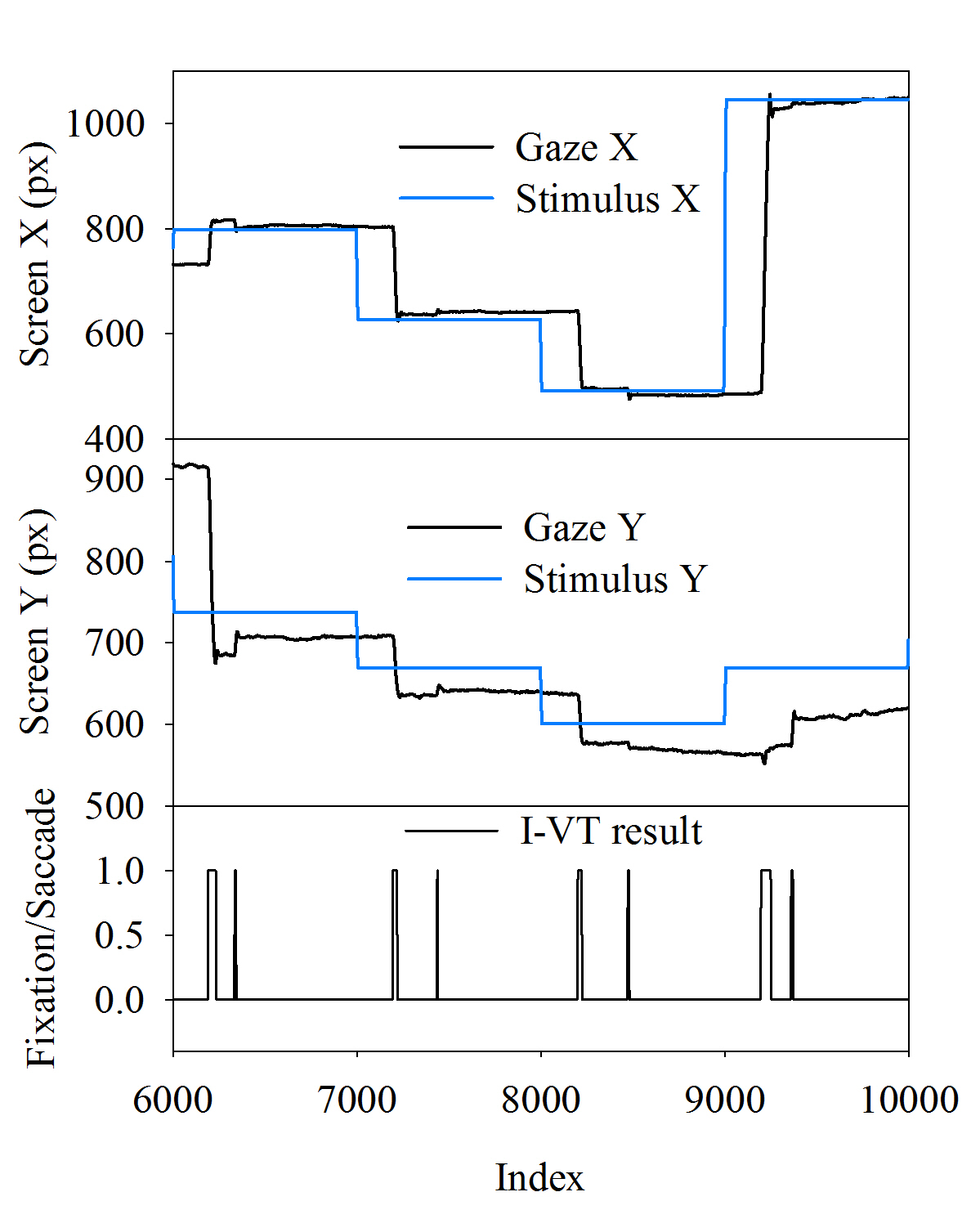}
\end{center}
\caption{Gaze data and stimulus for RAN\_30min sequence}
\label{fig:gazedata}
\label{fig:onecol}
\end{figure}
A minimum duration threshold of 100 milliseconds has been chosen to reduce the false positives in
fixation identification. Algorithm 1 returns the classification results for each data point as either fixation or saccade. Points that are not a part of fixations are considered as saccades in this stage. In the proposed approach, we consider saccades with their
durations more than a specified threshold to minimize the effect of spurious saccade segments. From the results of Algorithm 1, a list containing starting index and duration of all fixations and saccades is created. A post-processing stage is carried out to remove small-duration saccades. Saccades with duration less than 12 milliseconds are removed in this stage.

\subsubsection{Feature extraction}
After the removal of small-duration saccades, each eye movement data
is arranged into a sequence of fixations and saccades. The sequence of gaze locations and corresponding visual angles are also available for each fixation and saccade. Several statistical features are extracted from the position, velocity and acceleration profiles of
the gaze sequence. Other features like duration, dispersion, path length and co-occurrence features
are also extracted for both fixations and saccades. Earlier works ~\cite{komogortsev2010biometric} suggested that saccades provide a rich amount of information about the dynamics of oculomotor plant. Hence, we extract several other parameters including the saccadic ratio, main sequence, angle, etc. Saccades in horizontal and vertical directions are generated by different areas of the brain \cite{harwood2008optimally}. We use the statistical properties of the gaze data in $x$ and $y$ directions to incorporate this information. The distance and angle with the previous fixation/saccade are also used as features to leverage the temporal properties. The method used for computation of features is described below.

Let $X{\rm{ }} = {\rm{ }}\left\{ {{x_1},{x_2},{x_3},...,{x_N}} \right\}$
and $Y{\rm{ }} = {\rm{ }}\left\{ {{y_1},{y_2},{y_3},...,{y_N}} \right\}$ denote the set of coordinate positions of gaze in
each fixation/saccade and let $N$ denotes the number of data points in any fixation or saccade.
$\left( {{x_i},{y_i}} \right)$ denotes gaze location on the screen coordinate system and $\left( {\theta _i^x,\theta _i^y} \right)$ denotes the corresponding horizontal and vertical visual angles. 

A large number of features are extracted from the gaze sequence in each fixation and saccade. Some features are derived from the angular velocity. The differentiation operation for finding velocity and acceleration is carried out using forward difference method on the smoothed data. List of features extracted from fixations and saccades along with the methods of computation are shown in Table \ref{Tab:fixation_featuresused} and Table \ref{Tab:saccade_featuresused}. The features are extracted independently for each fixation and saccade.

\begin{table*}[htb]

\caption{\label{Tab:fixation_featuresused}List of features extracted from fixations}
\centering
\begin{tabular}{llll}
\hline
\multicolumn{2}{l}{Used in} & \multicolumn{1}{c}{Fixation Features} & Description \\ \cline{1-2}
TEX & RAN & & \\ \hline
N & Y & Fixation duration & Obtained from I-VT result \\
N & N & Standard Deviation (X) & \begin{tabular}[c]{@{}l@{}}From the screen coordinates\\ during fixation\end{tabular} \\
Y & N & Standard Deviation (Y) & '' \\
Y & Y & Path length & \begin{tabular}[c]{@{}l@{}}Length of path traveled in screen\\ $Path\,Length = \sum\limits_{i = 1}^{N - 1} {\sqrt {{{\left( {{x_{i + 1}} - {x_i}} \right)}^2} + {{\left( {{y_{i + 1}} - {y_i}} \right)}^2}} } $\end{tabular} \\
Y & Y & Angle with previous fixation & \begin{tabular}[c]{@{}l@{}}Angle with centroid of \\ previous fixation\end{tabular} \\
Y & Y & Distance from the last fixation & Euclidean distance from the previous fixation \\
Y & Y & Skewness(X) & From Screen coordinates \\
Y & Y & Skewness(Y) & '' \\
N & N & Kurtosis(X) & '' \\
Y & Y & Kurtosis(Y) & '' \\
Y & Y & Dispersion & \begin{tabular}[c]{@{}l@{}}Spatial spread during a fixation, Computed as\\$D= \left( {\max (X) - min(X)} \right) + \left( {\max (Y) - min(Y)} \right)$\end{tabular} \\
Y & Y & Average Velocity & $AV = Path\,Length/Duration$ \\ \hline
\multicolumn{4}{l}{$Y$ and $N$ denote inclusion or exclusion of the feature in the particular stimulus after feature selection}
\\ \hline
\end{tabular}

\end{table*}



\begin{table*}[!htb]
\caption{\label{Tab:saccade_featuresused}List of features extracted from saccades}
\centering
\begin{tabular}{llll}
\hline
\multicolumn{2}{l}{Used in} & \multicolumn{1}{c}{Saccade Features} & Description \\
TEX & RAN & & \\ \hline
N & N & Saccadic duration & Obtained from I-VT result \\
Y & Y & Dispersion & $D= \left( {\max (X) - min(X)} \right) + \left( {\max (Y) - min(Y)} \right)$, during saccade \\
NYYYYY & NNNYYY & M3S2K(Angular Velocity) & Features from angular velocity \\
YYYYYN & YYYYYY & M3S2K(Angular Acceleration) & Features from angular acceleration \\
Y & Y & Standard Deviation(X) & Obtained from screen positions \\
Y & Y & Standard Deviation(Y) & '' \\
Y & Y & Path length & \begin{tabular}[c]{@{}l@{}}Distance traveled in screen, \\$\sum\limits_{i = 1}^{N - 1} {\sqrt {{{\left( {{x_{i + 1}} - {x_i}} \right)}^2} + {{\left( {{y_{i + 1}} - {y_i}} \right)}^2}} }$\end{tabular} \\
Y & Y & Angle with previous saccade & \begin{tabular}[c]{@{}l@{}}Difference in saccadic angle with \\ previous saccade\end{tabular} \\
Y & Y & Distance from the previous saccade & \begin{tabular}[c]{@{}l@{}}Euclidean distance between\\ the centroid of the previous\\ saccade\end{tabular} \\

Y & Y & Saccadic ratio & $SR = \max (Angular\,Velocity)/Saccade\,Duration$ \\
Y & Y & Saccade angle & \begin{tabular}[c]{@{}l@{}}Obtained from fisrt and last points\\ as, $saccade\,angle = {\tan ^{ - 1}}\left( {\frac{{{y_N} - {y_1}}}{{{x_N} - {x_1}}}} \right)$\end{tabular} \\
Y & Y & Saccade amplitude &Obtained as: $\sqrt {{{\left( {{x_N} - {x_1}} \right)}^2} + {{\left( {{y_N} - {y_1}} \right)}^2}} $\\
YYYYYY & YYYYYY & M3S2K(Velocity\_X\_direction) & Features from screen positions \\
YYYYYY & YYYYNY & M3S2K(Velocity\_Y\_direction) & '' \\
YYYYYY & YYYYYY & M3S2K(Acceleration\_X\_direction) & '' \\
YYYYYY & YYNYYY & M3S2K(Acceleration\_Y\_direction) & '' \\
& & & \\ \hline
\multicolumn{4}{l}{*M3S2K - Statistical features:} \\
\multicolumn{4}{l}{Mean,Median,Max,Std, Skewness, Kurtosis}
\\ \hline
\multicolumn{4}{l}{$Y$ and $N$ denote inclusion or exclusion of the feature in the particular stimulus after feature selection} \\ \hline
\end{tabular}
\end{table*}


The control mechanisms generating fixations and saccades are different. The number of fixations and saccades is also different in each recording. There is a total of 12 and 46 features extracted from fixations and saccades respectively. A feature normalization scheme is used to scale each feature into a common range to ensure equal contribution in the final classification stage.

\subsubsection{Feature selection}

The large number of features extracted may contain redundancy and correlation. A backward feature selection algorithm, as shown in Algorithm 2 is used to retain a minimal set of discriminant features. We use the wrapper-based approach ~\cite{kohavi1997wrappers} for selecting the features. An RBFN classifier is used for finding the Equal Error Rate (EER) in each iteration. Cross-validation has been carried out in the training set to avoid overfitting. We used a random 50\% subset of the development dataset for the feature selection algorithm.
Feature selection algorithm starts with a set of all the features. Now in each iteration, the EER with inclusion and exclusion of a particular feature is found. The feature is retained if the EER with the use of the feature is better than EER with exclusion. The procedure is repeated for all the features in a sequential manner. The feature selection algorithm is
iterated ten times each time on a random 50\% subset for cross-validation. After these iterations, a set of important features is retained. To evaluate the generalization ability of the selected features, we have tested the algorithm (with the selected features) on an entirely disjoint set
that was not used in the feature selection process. The results with the evaluation set ~\cite{bioeye}(as shown in
the public results of BioEye 2015 competition) show the stability and generalization capability of the selected features. The subset of
features selected were different for different stimuli (TEX and RAN sets). The list of features selected for TEX and RAN stimuli is shown in Table \ref{Tab:fixation_featuresused} (Fixation features) and Table \ref{Tab:saccade_featuresused} (Saccade features). The features thus selected are used as inputs to the classification algorithm.


\begin{algorithm}[h]
\label{alg:backward}
\KwData{Feature matrix}
\KwResult{featureList[1: Included,0:Excluded]}
$ N\leftarrow$ Number of features\;
$ featureList\leftarrow ones(N)$\;
\For {$i \leftarrow$ 1 \textbf{to} N} {
$W \leftarrow featureList$\;
$E \leftarrow +Inf$\;
\For {$j \leftarrow$ 0 \textbf{to} 1} {
$ W[i] \leftarrow j$\;
T $\leftarrow$ EER with included features using RBFN\;
\If{$T < E$}{
$featureList[i]\leftarrow j$\;
$ E \leftarrow T$\;
}

}

}
\caption{Backward feature selection}
\end{algorithm}
After obtaining the set of features from fixations and saccades, we develop a model to represent the data. It has been empirically observed that the performance of classification approaches with Kernel-based methods are better than linear classifiers. It has also been reported that the parameters like amplitude-duration and amplitude-peak velocity may vary with the angle of saccade \cite{goossens1997human}. The nature of saccade dynamics may be different in different directions as the stimulus is changing randomly at various points on the screen. For each person, saccades of different amplitudes and directions form clusters in the feature space. In order to use the multi-mode nature of the data, we represent them by clustering them in the feature space. Representative vectors from each cluster are used to characterize each person. We use Gaussian Radial Basis Function Network (GRBFN) to model these data. The multiple cluster centers in the feature space are used as representative vectors in this approach. This vectors are selected using the K-means algorithm. Two different RBFNs are trained separately for fixation and saccade. Details about the structure of network and score fusion stage are described in the following section.
\subsection{RBF network} 
\label{sec:rbf}
Radial Basis Function Network (RBFN) is a class of neural networks initially proposed by Broomhead and Lowe \cite{broomhead1988radial}. Classification in RBFN is done by calculating the similarity between training and test vectors. Multiple prototype vectors corresponding to each class are stored in each neuron. The Euclidean distance between the input vector and the prototype vector is used to calculate neuron activations.

In the RBF network, input layer is made of feature vectors. $\varphi (x)$ is a radial basis function that finds the Euclidean distance between the input vector and the prototype vector. A weighted combination of scores from the RBF layer is used to classify the input into different categories.

The number of prototypes per class can be defined by the user, and these vectors can be found from the data using different algorithms like K-means, Linde-Buzo-Gray (LBG) algorithm, etc.

The Gaussian activation function of each neuron is chosen as:
\begin{equation}
\varphi (x) = {e^{ - \beta {{\left\| {x - \mu } \right\|}^2}}}
\end{equation}
where,
$\mu $ is the mean of the distribution. The parameter $\beta $ can be found from the data.

In this work, we have used K-means algorithm for selecting the representative vectors. For each
person, 32 clusters for fixations and 32 cluster centers for saccades are kept, resulting in $32N$
clusters for each RBFN (where $N$ is the number of persons in the dataset). The number of clusters
to keep is obtained empirically. We have clustered the fixations/saccades of each
individual separately to obtain a fixed number of representative vectors for each person.
A maximum of 100 iterations is used to form the clusters. A standard K-means algorithm is used with squared Euclidean distance, and the centers are updated in each
iteration. Each data point is assigned to the closest cluster center obtained from the K-means algorithm. For a particular neuron, the value of $\beta$ is computed from
the distance of all points belonging to that particular cluster as:
\begin{equation}
\beta = {1 \over {2{\sigma ^2}}}
\end{equation}
Where $\sigma$ is the mean Euclidean distance of the points (assigned to the specific neuron) from the centroid of the corresponding cluster.

\subsubsection{Notations}
The biometric identification problem is similar to a multiclass classification problem. Let there be $n$ samples of a $p$ dimensional data. Assume there are $m$ classes (corresponding to $m$ different individuals) with $c$ samples per class ($n = mc$). Let
${y_i}$ be the label corresponding to
${i^{th}}$ sample. Let $K$ be the number of representative vectors from each class. The value of $K$ is chosen empirically ($K=32$).

\subsubsection{Network learning}
The activations can be obtained as:
$A = {\varphi _{i,j}}({x_k}),\,\,i = 1,...,K,\,\,j = 1,...,m,\,\,k = 1,...,n$

The output of the network can be represented as a linear combination of the RBF activations as:

\begin{equation}
f(x) = \sum\limits_{j = 1}^m {{w_j}{\varphi _j}(x)}
\end{equation}

where,
$f(x)$ contains the class membership in vector form. Given the activations and output labels, the objective of the training stage is to find the weight parameters of the output layer. The weights are obtained by minimizing the sum of squared errors.

The output layer is represented by a linear system as:
\begin{equation}
A\hat w = \hat y
\end{equation}
The optimal set of weights can be found using the Moore-Penrose pseudoinverse. 
Alternatively, these weights can be learned through gradient descent method.
In the learning phase, features extracted from each fixation and saccade are used to train the model. Each fixation/saccade
is treated as a sample in the training process.

The method described here uses two-phase learning. RBF layer and weight layer trainings are carried out separately. However a joint training similar to back-propagation is also possible ~\cite{schwenker2001three}.

\subsubsection{Training stage}
Only the session 1 data from the datasets are used in the training stage. Cluster centers and corresponding $\beta$ values are computed separately for each person (resulting in $32N$ neurons for both fixation and saccade RBFNs). The output weights (${\hat w_{fix}}$ and ${\hat w_{sacc}}$) are found using all fixations and saccades from all the subjects in the dataset.
\begin{figure*}[htb]
\centering
\includegraphics[width=0.8\linewidth]{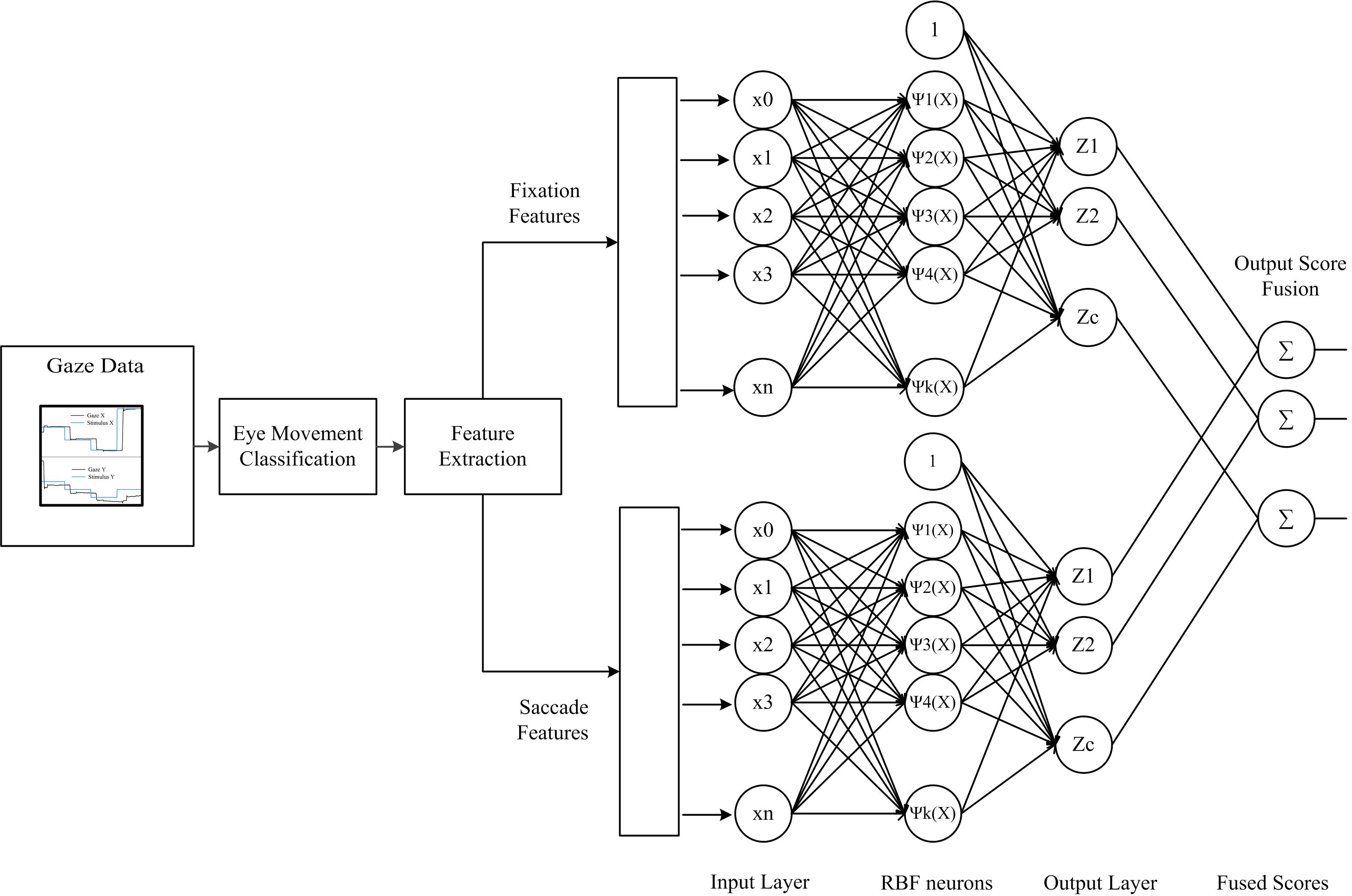}
\caption{Schematic of the proposed framework.}
\label{fig:rbfncombined}
\end{figure*}
\subsubsection{Testing stage}
Session 2 data is used in the testing stage. Parameters of RBFN are computed separately for fixations and saccades in the training session. The scores from both RBFNs are combined to obtain the final result. The overall configuration of the scheme is shown in Fig. \ref{fig:rbfncombined}.

 For an unlabeled probe, the activations for each fixation and saccade (${A_{fix}}$ and ${A_{sacc}}$) are found separately using the cluster centers obtained in the training stage.
The final classification is carried out using the combined score obtained from all saccades and
fixations. 
 Let $n{}_{fix}$ and $n{}_{sacc}$ be the number of fixations and saccades in an unlabeled gaze sequence.
The combined score can be obtained as:
\begin{equation}
score = \lambda {1 \over {{n_{fix}}}}\sum\limits_{i = 1}^{{n_{fix}}} {A_{fix}^i{{\hat w}_{fix}} + \left( {1 - \lambda } \right){1 \over {{n_{sacc}}}}} \sum\limits_{i = 1}^{{n_{sacc}}} {A_{sacc}^i{{\hat w}_{sacc}}}
\end{equation}

where, $\lambda \in \left[ {0\,1} \right]$ is the weight used in the score fusion.
The parameter $\lambda$ decides the contribution of fixations and saccades in the final decision stage.
This value can be obtained empirically. In the present work, $\lambda$ value of 0.5 is
used.

The label of the unknown sample can be obtained as:
\begin{equation}
label = \mathop {\arg \max }\limits_m (score)
\end{equation}

\section{Experiments and results}
\begin{table*}[!htb]
\caption{\label{datasetdetails} Details about the database}
\centering
\begin{tabular}{@{}lllll@{}}
\toprule
\textbf{Dataset Name} & \textbf{RAN\_30min} & \textbf{RAN\_1year} & \textbf{TEX\_30min} & \textbf{TEX\_1year} \\ \midrule
Number of subjects & 153 & 37 & 153 & 37 \\ \midrule
Stimulus & \begin{tabular}[c]{@{}l@{}}White dot moving\\ in a dark\\  background\end{tabular} & \begin{tabular}[c]{@{}l@{}}White dot moving \\ in a dark\\ background\end{tabular} & \begin{tabular}[c]{@{}l@{}}Text \\ excerpt\end{tabular} & \begin{tabular}[c]{@{}l@{}}Text \\ excerpt\end{tabular} \\ \midrule
Duration of experiment & 100 seconds & 100 seconds & 60 seconds & 60 seconds \\ \midrule
\begin{tabular}[c]{@{}l@{}}Interval between\\ training\\ and testing data\end{tabular} & 30 minutes & 1 year & 30 minutes & 1 year \\ \bottomrule
\end{tabular}
\end{table*}

\subsection{Datasets}
The data used in this paper are part of the development phase of BioEye 2015 ~\cite{bioeye} competition. Data recorded in three different sessions are available. First two sessions are separated by a time interval of 30 minutes containing recordings of 153 subjects (ages 18-43). A third session, conducted after 1 year, (37 subjects) is also available to evaluate the robustness against template aging. The database contains gaze sequences obtained using two distinct types of visual stimuli. In one set (RAN), a white dot moving in a dark background was used as the stimulus. The subjects were asked to follow the dot. Text excerpt shown on the screen was used as the stimulus in the other set (TEX). The samples were recorded with an EyeLink eye-tracker (with a reported spatial accuracy of 0.5 degrees) at 1000 Hz and down-sampled to 250 Hz with anti-aliasing filtering. The development dataset contains the ground truth about the identity of the persons. An additional evaluation set is also available without ground truth. 

In each recording, visual angles in $x$ and $y$ direction, stimulus angle in $x$ and $y$ direction and information regarding the validity of the samples are available. Details about the stimulus types in BioEye2015 database are given below.
\subsubsection{Random dot stimulus (RAN\_30min \& RAN\_1year)}
The stimulus used was a white dot appearing at random locations on a black computer screen. The position of the stimulus would change every second. The subjects were asked to follow the dot on the screen and recording was carried out for 100 seconds.
\subsubsection{Text stimulus (TEX\_30min \& TEX\_1year)}
The task, in this case, was reading text excerpts from the poem of Lewis Carroll ``The Hunting of the Snark''. The duration of this experiment was 60 seconds.

 A comprehensive list of the datasets and parameters are shown in Table \ref{datasetdetails}.
\subsection{Evaluation metrics}
The proposed algorithm has been evaluated in the labeled development set. Rank-1 accuracy and EER are used for evaluating the algorithm. Rank-1 (R1) accuracy is defined as the ratio of the total number of correct detections to the number of samples used. EER is the percentage at which False Acceptance Rate (FAR) and False Rejection Rate (FRR) are equal. Detection Error Trade-off (DET) curves are shown for all the datasets. Rank(\textit{n}) accuracy is the number of correct detections in the top $n$ candidates. Cumulative match characteristics (CMC) is the cumulative plot of rank(n) accuracy. CMC curves are also plotted for all the four datasets. The evaluation set in the BioEye2015 dataset is unlabeled. However, we report the R1 accuracy as obtained from the public results ~\cite{bioeye} of the competition.
\subsection{Results}
\subsubsection{Performance in the development datasets}
The model was trained using 50\% of data in the development datasets.
We have trained and tested the algorithm on completely disjoint sessions to test its generalization ability. For example, in RAN\_30min
sequence there are 153 samples available for two different sessions. We have trained the
Algorithm only on the first session (using a random 50\% subset of the data). The evaluation was
carried out on the session 2 data. We have not used the data from the same session for training
and testing since it won't account for intersession variability. 

The average R1 accuracy and EER were calculated from random 50\% subsets of development datasets. This procedure was repeated 100 times and the average R1 accuracy and EER were
obtained. The results obtained along with the standard deviations are given in Table \ref{resultsdev}.
\begin{figure*}[!htb]
\centering
\includegraphics[width=0.8\linewidth]{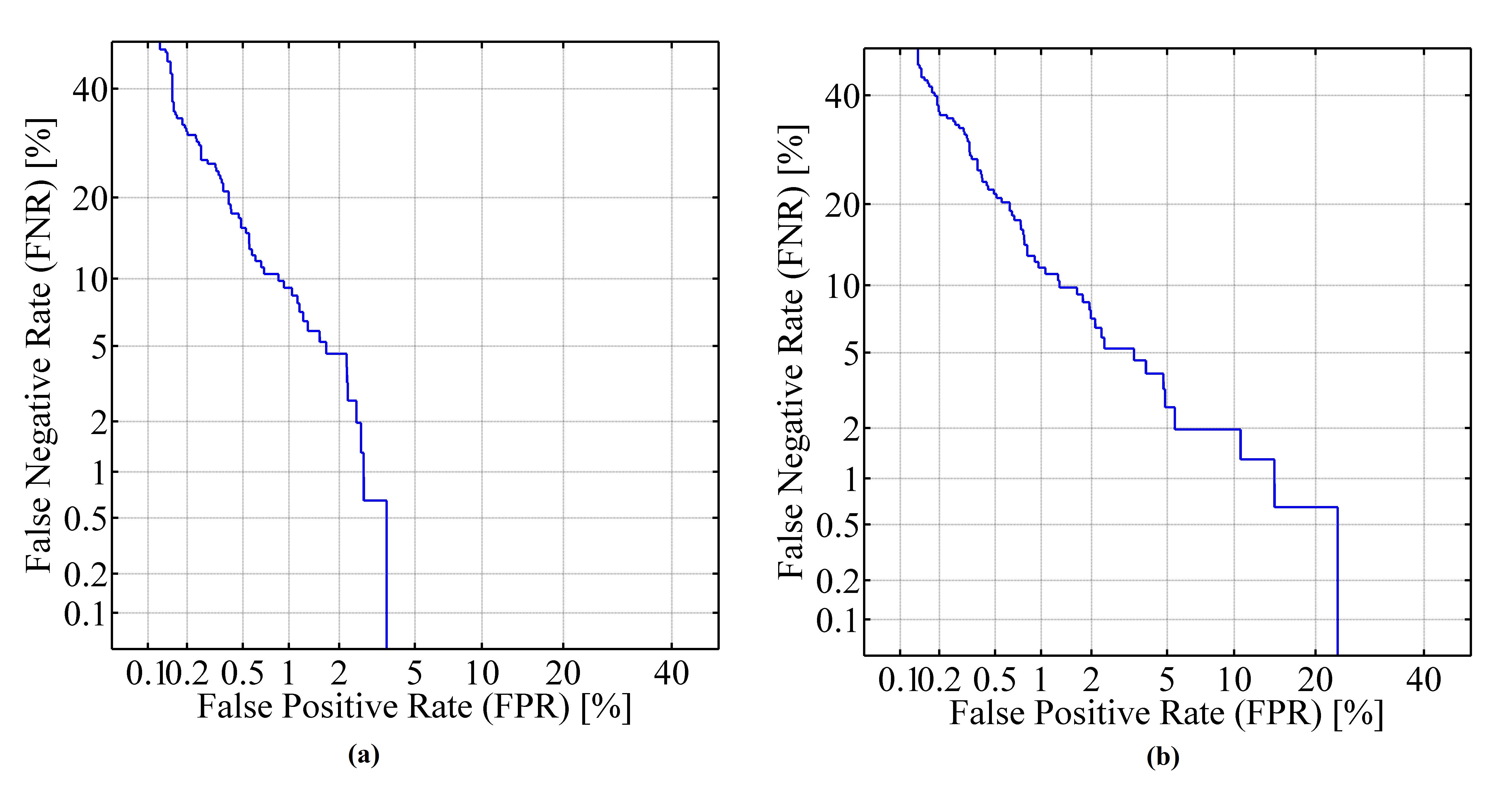}
\caption{DET curve for (a) RAN\_30min and (b) TEX\_30min}
\label{fig:eer_30_dev}
\end{figure*}
\begin{figure*}[!htb]
\centering
\includegraphics[width=0.8\linewidth]{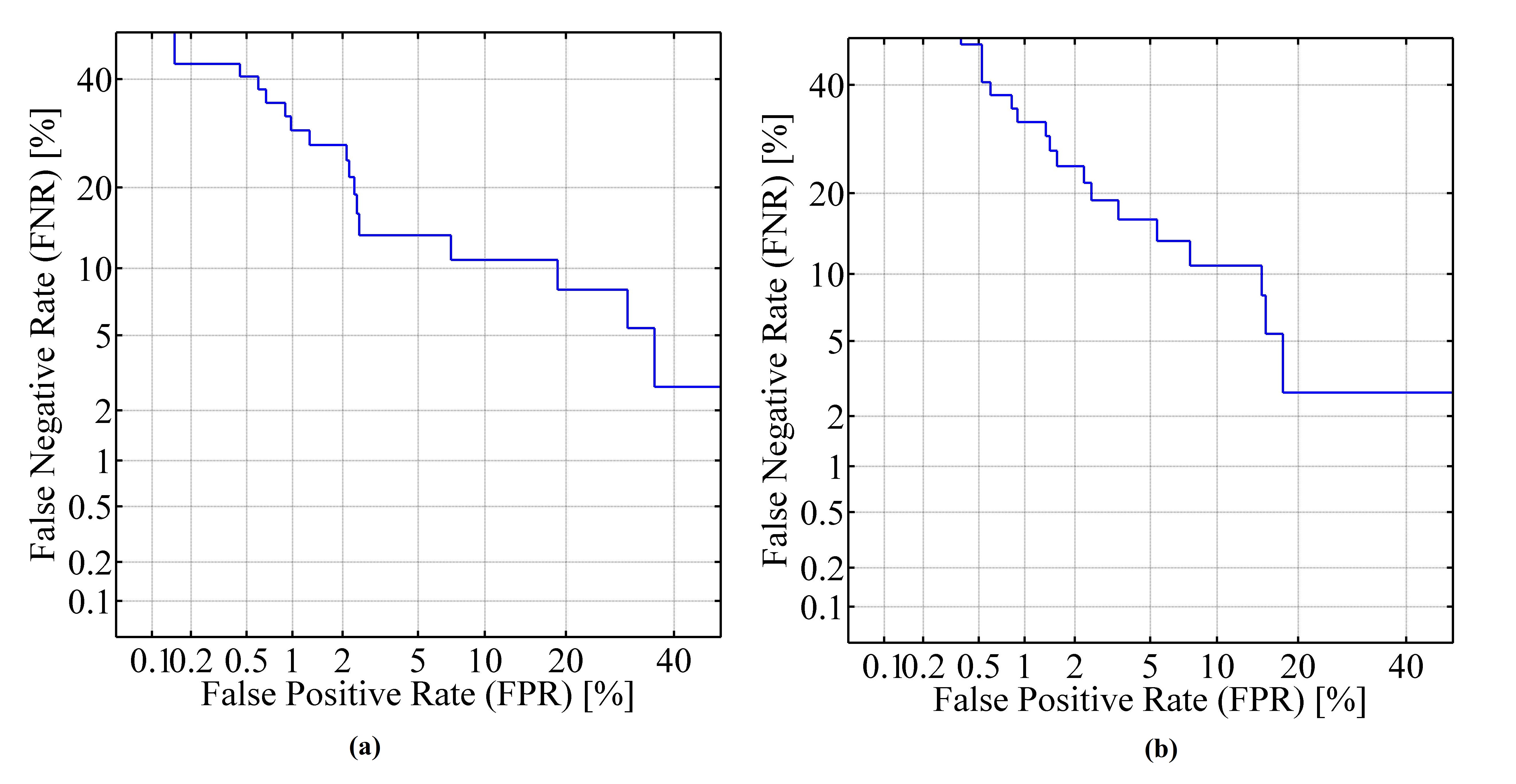}
\caption{DET curve for (a) RAN\_1year and (b) TEX\_1year}
\label{fig:eer_1yr_dev}
\end{figure*}

The R1 accuracy in RAN\_30min and TEX\_30min databases are above 90\% indicating the robustness of the proposed framework. The EER on RAN\_30min database is found out to be 2.59\%, comparable to the accuracy levels of fingerprint (2.07\% EER) \cite{maio2004fvc2004}, voice recognition systems, and facial geometry (15\% EER) ~\cite{phillips2010frvt} biometrics.
\begin{table}[h]
\centering
\caption{\label{resultsdev} Results in the development datasets}
\begin{tabular}{@{}lllll@{}}
\toprule
& RAN\_30 & RAN\_1yr & TEX\_30 & TEX\_1yr \\ \midrule
R1 & 90.10$\pm$2.76 & 79.31$\pm$6.86 & 92.38$\pm$2.56 & 83.41$\pm$6.98 \\
EER & 2.59$\pm$0.71 & 10.96$\pm$4.59 & 3.78$\pm$0.77 & 9.36$\pm$3.49 \\ \bottomrule
\end{tabular}
\end{table}

\begin{figure*}[!htb]
\centering
\includegraphics[width=0.9\linewidth]{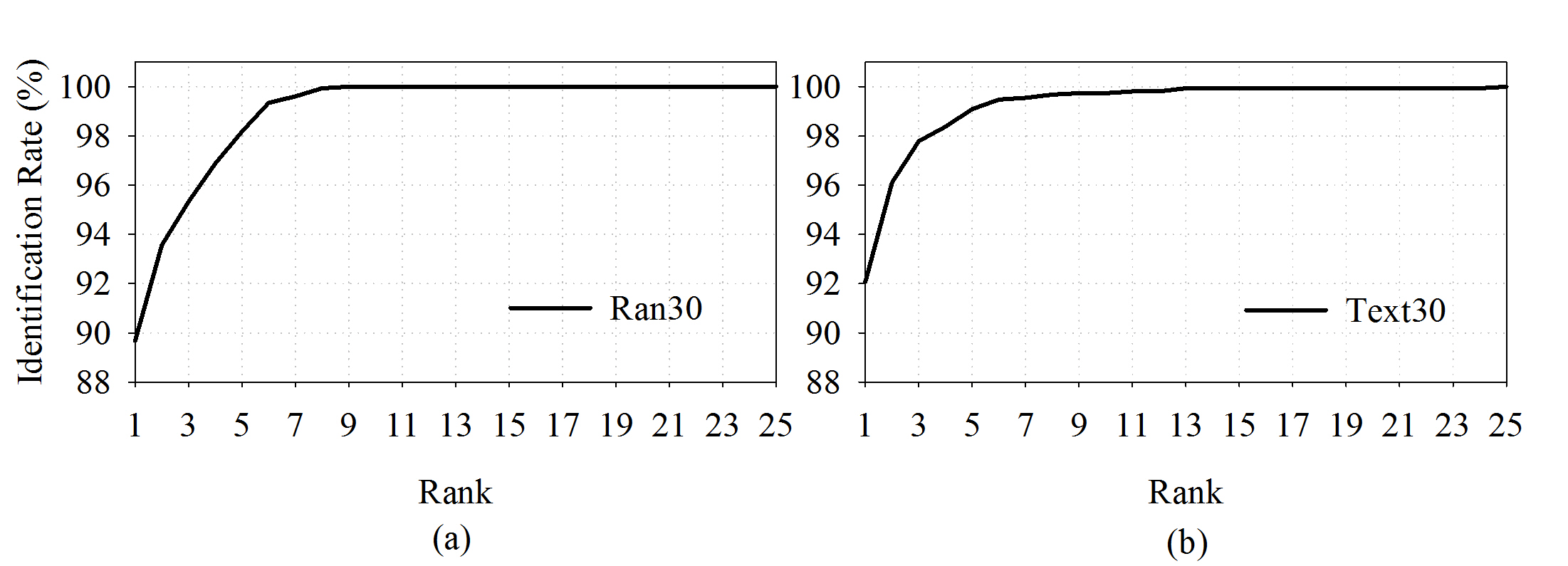}
\caption{CMC curve for (a) RAN\_30min and (b) TEX\_30min}
\label{fig:cmc_30_dev}
\end{figure*}

\begin{figure*}[!htb]
\centering
\includegraphics[width=0.9\linewidth]{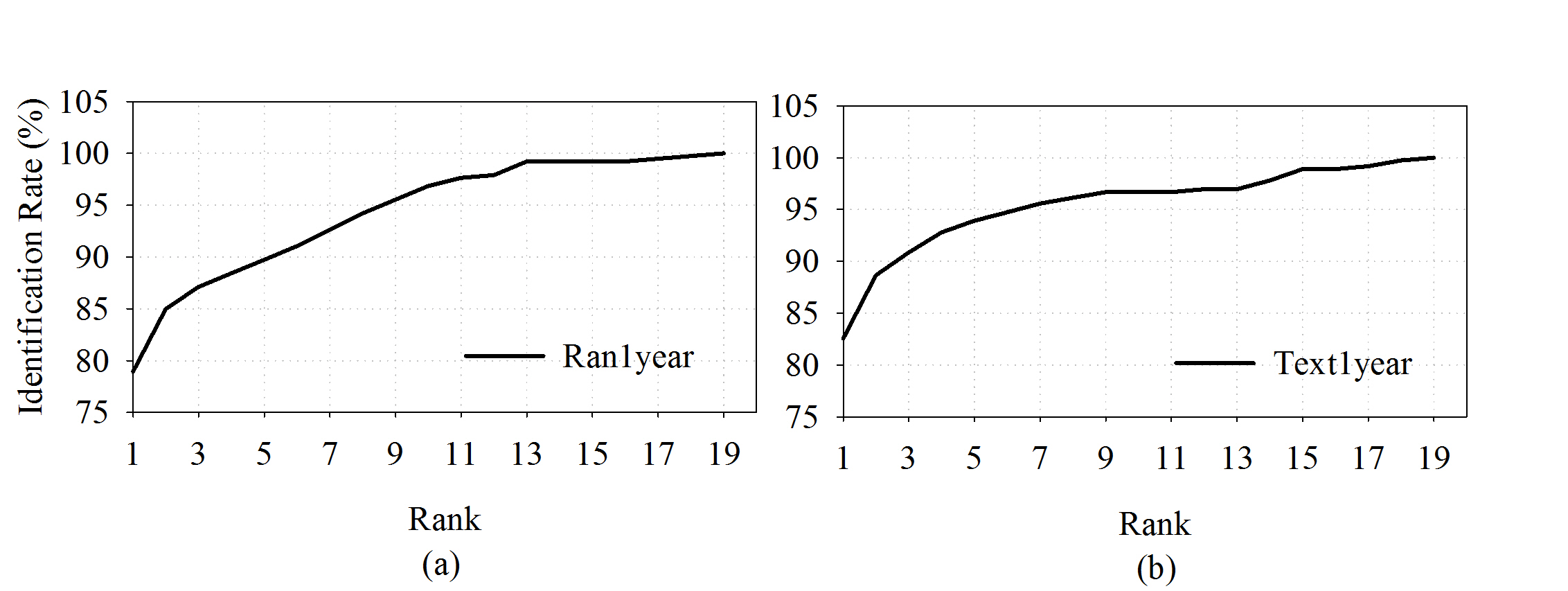}
\caption{CMC curve for (a) RAN\_1year and (b) TEX\_1year}
\label{fig:cmc_1yr_dev}
\end{figure*}

R1 accuracy (Table \ref{comparedev}) of the proposed algorithm obtained from the development set was compared with the baseline algorithm (CEM-B) \cite{holland2013complexa}.
The average cumulative matching characteristics curves for the four datasets are shown in Fig. \ref{fig:cmc_30_dev} and Fig. \ref{fig:cmc_1yr_dev}.

 The Detection Error Trade-off (DET) curves for the development datasets are shown in Fig. \ref{fig:eer_30_dev} and Fig. \ref{fig:eer_1yr_dev}. In Fig. \ref{fig:eer_30_dev} (a) and (b), FNR becomes very small as FPR increases indicating a good separation from
impostors. The reduction in FNR may be because of the addition of scores of all the fixations and saccades
in the score fusion stage. Impostor scores are considerably smaller than genuine scores in the proposed approach. The performance in
1-year sessions is poor compared to 30-minutes sessions indicating template aging effects.

\begin{table}[h]
\centering
\caption{\label{comparedev} Comparison of R1 accuracy in the entire development dataset}
\begin{tabular}{@{}lllll@{}}
\toprule
& RAN\_30 & RAN\_1yr & TEX\_30 & TEX\_1yr \\ \midrule
Our Method & 89.54\% & 81.08\% & 85.62\% & 78.38\% \\
\begin{tabular}[c]{@{}l@{}} Baseline \end{tabular} & 40.52\% & 16.22\% & 52.94\% & 40.54\% \\ \bottomrule
\end{tabular}
\end{table}

\subsubsection{Performance in the evaluation sets}
The evaluation part of the database is unlabeled. However, the results of the competition are available on the website ~\cite{bioeye}.  The evaluation set of the dataset had only one unlabeled data for every labeled sample. We have used this one to one correspondence assumption in the final stage of the algorithm.

Let there be $n$ labeled and $n$ unlabeled recordings. The task is to assign each unlabeled file to a labeled file.  The scores obtained from RBF output stage were stored in a matrix $D$ (with dimension $n$x$n$).  $D(i,j)$ denotes the normalized similarity score  between ${i^{th}}$ labeled and ${j^{th}}$ unlabeled samples. We have selected the best match for each unlabeled recording using Algorithm 3.  The use of this one to one assumption improved the results. However, this assumption may not be suitable for practical biometric identification/verification scenarios. The proposed method has been found to outperform all the other methods even without the one to one assumption indicating the robustness for biometric applications. The results with and without this assumption are shown in Table \ref{compareeval}. 
\begin{algorithm}[h]
\label{alg:onetoone}
\KwData{$D$ (Score matrix)}
\KwResult{Matches}
$[n,n]=size(D)$\;


\For {$i \leftarrow$ 1 \textbf{to} n} {
$[row,col]=find(D==max(D(:)))$\;
$D(row,:)=-\infty $\;
$D(,:col)=-\infty $\;
pair=$[row,col]$\;
Matches.\textit{append}(pair)\;

}
\caption{One to one matching}
\end{algorithm}
\begin{table}[h]
\centering

\caption{\label{compareeval} Comparison of R1 accuracy with baseline method in evaluation dataset}
\label{my-label}
\begin{tabular}{@{}lllll@{}}

\toprule
& RAN\_30 & RAN\_1yr & TEX\_30 & TEX\_1yr \\ \midrule
\begin{tabular}[c]{@{}l@{}}Our Method \end{tabular} & 93.46\% & 83.78\% & 89.54\% & 83.78\% \\
\begin{tabular}[c]{@{}l@{}}Our Method* \end{tabular} & 98.69\% & 89.19\% & 98.04\% & 94.59\% \\
Baseline & 33.99\% & 40.54\% & 58.17\% & 48.65\% \\ \bottomrule
\end{tabular}
\end{table}

\subsection{Computational complexity}
The algorithm has been implemented in an Intel Core i5 CPU, 3.33 GHz PC with 4 GB RAM. The average training time for the network without code optimization (single-threaded) in MATLAB is about 400 seconds (with 153 samples). In the testing phase, for predicting one unlabeled recording, it takes on an average 0.21 seconds (in TEX\_30min). The time taken for training and testing phase can be improved considerably by implementation in C, C++, using parallel processing platforms like Graphical Processing Units (GPU).
\subsection{ Discussions}
\subsubsection{Performance of the algorithm}
The R1 accuracy of the proposed method is high in both TEX and RAN datasets, which indicates the possibility of developing a task-independent biometric system. The EER and R1 accuracy achieved show the robustness of the proposed score fusion approach. The selected features show good discrimination ability in both stimuli. The accuracy with 1-year datasets is comparatively lesser than that with the 30-minute datasets.
 This lower accuracy may be attributed to template aging effects.  
 Some of the selected features may show variability over time ~\cite{komogortsev2014template} ~\cite{kasprowski2013impact}. 

The feature selection was carried out in 30-minute datasets due to the availability of a large number of subjects.  Feature selection with 1-year datasets may lead to overfitting because of fewer subjects. This issue can be solved by using the feature selection in 1-year datasets with a larger number of subjects, which may identify features that are robust against template aging.  However, the results show significant improvement compared to the state of the art methods. The proposed algorithm was ranked first in the BioEye 2015 ~\cite{bioeye} competition.
\subsubsection{Limitations}
Controlled experimental setup was used to collect the data used in this work. The sampling rate and quality of data used in the present work were very high since it was collected in lab conditions using chinrest. Accurate estimation of the features in noisy, low sampling rates is necessary for the use in a practical biometric scenario. The nature of eye movements may be affected by the level of alertness, fatigue, emotions, cognitive loading, etc. Consumption of caffeine and alcohol by the subjects may affect the performance of the proposed algorithm. The features selected for biometrics should be invariant to such variations. Only two sessions of data were available for each subject. Intersession variability and template aging effects need to be studied further. Lack of publicly available databases containing a large number of samples (to account for template aging, uncontrolled environment, affective states, intersession variability) is another problem. Creation of a large database with such variability could provide more robust solutions.

\section{Conclusions}
This work proposes a novel framework for biometric identification based on dynamic characteristics of eye movements. The raw eye movement data is classified into a sequence of fixations and saccades. We extract a large set of features from fixations and saccades to characterize each individual. The important features extracted from fixations and saccades are identified based on a backward selection framework. Two different Gaussian RBF networks are trained using features from fixations and saccades separately. In the detection phase, scores obtained from both RBF networks are used to get the subject's identity. The high accuracy obtained shows the robustness of the proposed algorithm. The proposed framework can be easily integrated into the existing iris recognition systems. A combination of the proposed approach with conventional iris recognition systems may give rise to a new counterfeit-resistant biometric system. The comparable accuracy in distinct types of stimuli indicates the possibility of developing a task-independent system for eye movement biometrics. The proposed method can also be used for continuous authentication in desktop environments. Robustness of the algorithm against lower sampling rates, calibration error and noise can be explored in future. The effect of duration of data on the level of accuracy is another topic to be investigated. 

\ifCLASSOPTIONcompsoc

  \section*{Acknowledgments}
\else

  \section*{Acknowledgment}
\fi

The authors would like to thank the organizers of BioEye 2015 competition for providing the data.

\ifCLASSOPTIONcaptionsoff
  \newpage
\fi

\bibliographystyle{myIEEEtran}
\bibliography{refsnew}

\begin{thebibliography}{10}
\providecommand{\url}[1]{#1}
\csname url@samestyle\endcsname
\providecommand{\newblock}{\relax}
\providecommand{\bibinfo}[2]{#2}
\providecommand{\BIBentrySTDinterwordspacing}{\spaceskip=0pt\relax}
\providecommand{\BIBentryALTinterwordstretchfactor}{4}
\providecommand{\BIBentryALTinterwordspacing}{\spaceskip=\fontdimen2\font plus
\BIBentryALTinterwordstretchfactor\fontdimen3\font minus
  \fontdimen4\font\relax}
\providecommand{\BIBforeignlanguage}[2]{{%
\expandafter\ifx\csname l@#1\endcsname\relax
\typeout{** WARNING: IEEEtran.bst: No hyphenation pattern has been}%
\typeout{** loaded for the language `#1'. Using the pattern for}%
\typeout{** the default language instead.}%
\else
\language=\csname l@#1\endcsname
\fi
#2}}
\providecommand{\BIBdecl}{\relax}
\BIBdecl

\bibitem{jain2007handbook}
A.~K. Jain, P.~Flynn, and A.~A. Ross, \emph{Handbook of biometrics}.\hskip 1em
  plus 0.5em minus 0.4em\relax Springer Science \& Business Media, 2007.

\bibitem{jain2004introduction}
A.~K. Jain, A.~Ross, and S.~Prabhakar, ``An introduction to biometric
  recognition,'' \emph{Circuits and Systems for Video Technology, IEEE
  Transactions on}, vol.~14, no.~1, pp. 4--20, 2004.

\bibitem{wang2009behavioral}
L.~Wang, X.~Geng, L.~Wang, and X.~Geng, \emph{Behavioral Biometrics For Human
  Identification: Intelligent Applications}.\hskip 1em plus 0.5em minus
  0.4em\relax IGI Global, 2009.

\bibitem{marcel2007person}
S.~Marcel and J.~d.~R. Mill{\'a}n, ``Person authentication using brainwaves
  (eeg) and maximum a posteriori model adaptation,'' \emph{Pattern Analysis and
  Machine Intelligence, IEEE Transactions on}, vol.~29, no.~4, pp. 743--752,
  2007.

\bibitem{plataniotis2006ecg}
K.~N. Plataniotis, D.~Hatzinakos, and J.~K. Lee, ``Ecg biometric recognition
  without fiducial detection,'' in \emph{Biometric Consortium Conference, 2006
  Biometrics Symposium: Special Session on Research at the}.\hskip 1em plus
  0.5em minus 0.4em\relax IEEE, 2006, pp. 1--6.

\bibitem{ib2005independent}
{I.B. Group}, ``Independent testing of iris recognition technology,''
  \emph{Final Report, NBCHC030114/0002}, 2005.

\bibitem{roberts2007biometric}
C.~Roberts, ``Biometric attack vectors and defences,'' \emph{Computers \&
  Security}, vol.~26, no.~1, pp. 14--25, 2007.

\bibitem{schuckers2002issues}
S.~Schuckers, L.~Hornak, T.~Norman, R.~Derakhshani, and S.~Parthasaradhi,
  ``Issues for liveness detection in biometrics,'' in \emph{Proceedings of
  Biometric Consortium Conference. IEEE, New York}, 2002.

\bibitem{leigh1999neurology}
R.~J. Leigh and D.~S. Zee, \emph{The neurology of eye movements}.\hskip 1em
  plus 0.5em minus 0.4em\relax Oxford university press New York, 1999, vol.~90.

\bibitem{kinnunen2010towards}
T.~Kinnunen, F.~Sedlak, and R.~Bednarik, ``Towards task-independent person
  authentication using eye movement signals,'' in \emph{Proceedings of the 2010
  Symposium on Eye-Tracking Research \& Applications}.\hskip 1em plus 0.5em
  minus 0.4em\relax ACM, 2010, pp. 187--190.

\bibitem{kasprowski2004eye}
P.~Kasprowski and J.~Ober, ``Eye movements in biometrics,'' in \emph{Biometric
  Authentication}.\hskip 1em plus 0.5em minus 0.4em\relax Springer, 2004, pp.
  248--258.

\bibitem{bednarik2005eye}
R.~Bednarik, T.~Kinnunen, A.~Mihaila, and P.~Fr{\"a}nti, ``Eye-movements as a
  biometric,'' in \emph{Image analysis}.\hskip 1em plus 0.5em minus 0.4em\relax
  Springer, 2005, pp. 780--789.

\bibitem{komogortsev2010biometric}
O.~V. Komogortsev, S.~Jayarathna, C.~R. Aragon, and M.~Mahmoud, ``Biometric
  identification via an oculomotor plant mathematical model,'' in
  \emph{Proceedings of the 2010 Symposium on Eye-Tracking Research \&
  Applications}.\hskip 1em plus 0.5em minus 0.4em\relax ACM, 2010, pp. 57--60.

\bibitem{komogortsev2012biometric}
O.~V. Komogortsev, A.~Karpov, L.~R. Price, and C.~Aragon, ``Biometric
  authentication via oculomotor plant characteristics,'' in \emph{Biometrics
  (ICB), 2012 5th IAPR International Conference on}.\hskip 1em plus 0.5em minus
  0.4em\relax IEEE, 2012, pp. 413--420.

\bibitem{holland2013complexb}
C.~D. Holland and O.~V. Komogortsev, ``Complex eye movement pattern biometrics:
  the effects of environment and stimulus,'' \emph{Information Forensics and
  Security, IEEE Transactions on}, vol.~8, no.~12, pp. 2115--2126, 2013.

\bibitem{rigas2012biometric}
I.~Rigas, G.~Economou, and S.~Fotopoulos, ``Biometric identification based on
  the eye movements and graph matching techniques,'' \emph{Pattern Recognition
  Letters}, vol.~33, no.~6, pp. 786--792, 2012.

\bibitem{rigas2012human}
I.~Rigas, G.~Economou, and S.~Fotopoulos, ``Human eye movements as a trait for
  biometrical identification,'' in \emph{Biometrics: Theory, Applications and
  Systems (BTAS), 2012 IEEE Fifth International Conference on}.\hskip 1em plus
  0.5em minus 0.4em\relax IEEE, 2012, pp. 217--222.

\bibitem{zhang2012biometric}
Y.~Zhang and M.~Juhola, ``On biometric verification of a user by means of eye
  movement data mining,'' in \emph{IMMM 2012, The Second International
  Conference on Advances in Information Mining and Management}, 2012, pp.
  85--90.

\bibitem{cantoni2015gant}
V.~Cantoni, C.~Galdi, M.~Nappi, M.~Porta, and D.~Riccio, ``Gant: Gaze analysis
  technique for human identification,'' \emph{Pattern Recognition}, vol.~48,
  no.~4, pp. 1027--1038, 2015.

\bibitem{holland2011biometric}
C.~Holland and O.~V. Komogortsev, ``Biometric identification via eye movement
  scanpaths in reading,'' in \emph{Biometrics (IJCB), 2011 International Joint
  Conference on}.\hskip 1em plus 0.5em minus 0.4em\relax IEEE, 2011, pp. 1--8.

\bibitem{holland2013complexa}
C.~D. Holland and O.~V. Komogortsev, ``Complex eye movement pattern biometrics:
  Analyzing fixations and saccades,'' in \emph{Biometrics (ICB), 2013
  International Conference on}.\hskip 1em plus 0.5em minus 0.4em\relax IEEE,
  2013, pp. 1--8.

\bibitem{bioeye}
``Bioeye2015,competition on biometrics via eye movements,''
  \url{http://bioeye.cs.txstate.edu/}, accessed: 2015-04-09.

\bibitem{collewijn1988binocular}
H.~Collewijn, C.~J. Erkelens, and R.~Steinman, ``Binocular co-ordination of
  human horizontal saccadic eye movements.'' \emph{The Journal of Physiology},
  vol. 404, no.~1, pp. 157--182, 1988.

\bibitem{harris1984instrument}
C.~M. Harris, I.~Abramov, and L.~Hainl, ``Instrument considerations in
  measuring fast eye movements,'' \emph{Behavior Research Methods, Instruments,
  \& Computers}, vol.~16, no.~4, pp. 341--350, 1984.

\bibitem{krishnan2013selection}
S.~R. Krishnan and C.~S. Seelamantula, ``On the selection of optimum
  savitzky-golay filters,'' \emph{Signal Processing, IEEE Transactions on},
  vol.~61, no.~2, pp. 380--391, 2013.

\bibitem{savitzky1964smoothing}
A.~Savitzky and M.~J. Golay, ``Smoothing and differentiation of data by
  simplified least squares procedures.'' \emph{Analytical chemistry}, vol.~36,
  no.~8, pp. 1627--1639, 1964.

\bibitem{holland2012biometric}
C.~D. Holland and O.~V. Komogortsev, ``Biometric verification via complex eye
  movements: The effects of environment and stimulus,'' in \emph{Biometrics:
  Theory, Applications and Systems (BTAS), 2012 IEEE Fifth International
  Conference on}.\hskip 1em plus 0.5em minus 0.4em\relax IEEE, 2012, pp.
  39--46.

\bibitem{salvucci2000identifying}
D.~D. Salvucci and J.~H. Goldberg, ``Identifying fixations and saccades in
  eye-tracking protocols,'' in \emph{Proceedings of the 2000 symposium on Eye
  tracking research \& applications}.\hskip 1em plus 0.5em minus 0.4em\relax
  ACM, 2000, pp. 71--78.

\bibitem{harwood2008optimally}
M.~R. Harwood and J.~P. Herman, ``Optimally straight and optimally curved
  saccades,'' \emph{The Journal of Neuroscience}, vol.~28, no.~30, pp.
  7455--7457, 2008.

\bibitem{kohavi1997wrappers}
R.~Kohavi and G.~H. John, ``Wrappers for feature subset selection,''
  \emph{Artificial intelligence}, vol.~97, no.~1, pp. 273--324, 1997.

\bibitem{goossens1997human}
H.~H. Goossens and A.~Van~Opstal, ``Human eye-head coordination in two
  dimensions under different sensorimotor conditions,'' \emph{Experimental
  Brain Research}, vol. 114, no.~3, pp. 542--560, 1997.

\bibitem{broomhead1988radial}
D.~S. Broomhead and D.~Lowe, ``Radial basis functions, multi-variable
  functional interpolation and adaptive networks,'' DTIC Document, Tech. Rep.,
  1988.

\bibitem{schwenker2001three}
F.~Schwenker, H.~A. Kestler, and G.~Palm, ``Three learning phases for
  radial-basis-function networks,'' \emph{Neural networks}, vol.~14, no.~4, pp.
  439--458, 2001.

\bibitem{maio2004fvc2004}
D.~Maio, D.~Maltoni, R.~Cappelli, J.~L. Wayman, and A.~K. Jain, ``Fvc2004:
  Third fingerprint verification competition,'' in \emph{Biometric
  Authentication}.\hskip 1em plus 0.5em minus 0.4em\relax Springer, 2004, pp.
  1--7.

\bibitem{phillips2010frvt}
P.~J. Phillips, W.~T. Scruggs, A.~J. O'Toole, P.~J. Flynn, K.~W. Bowyer, C.~L.
  Schott, and M.~Sharpe, ``Frvt 2006 and ice 2006 large-scale experimental
  results,'' \emph{Pattern Analysis and Machine Intelligence, IEEE Transactions
  on}, vol.~32, no.~5, pp. 831--846, 2010.

\bibitem{komogortsev2014template}
O.~V. Komogortsev, C.~D. Holland, and A.~Karpov, ``Template aging in eye
  movement-driven biometrics,'' in \emph{SPIE Defense+ Security}.\hskip 1em
  plus 0.5em minus 0.4em\relax International Society for Optics and Photonics,
  2014, pp. 90\,750A--90\,750A.

\bibitem{kasprowski2013impact}
P.~Kasprowski, ``The impact of temporal proximity between samples on eye
  movement biometric identification,'' in \emph{Computer Information Systems
  and Industrial Management}.\hskip 1em plus 0.5em minus 0.4em\relax Springer,
  2013, pp. 77--87.

\end{thebibliography}
\begin{IEEEbiography}[{\includegraphics[width=1in,height=1.25in,clip,keepaspectratio]{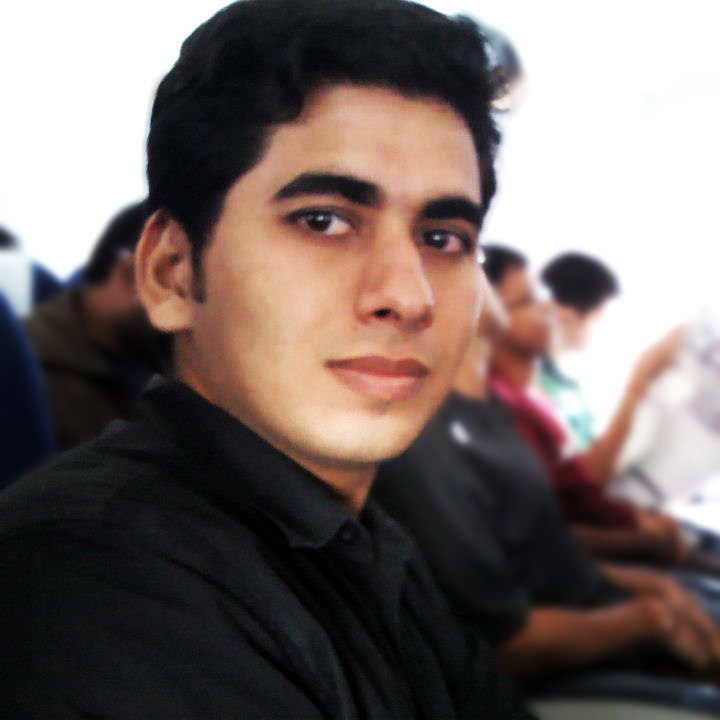}}]{Anjith George}
has received his B.Tech. (Hons.)
degree from Calicut University, India in 2010. and
M-Tech degree from Indian Institute of Technology
(IIT) Kharagpur, India in 2012. Presently he is
pursuing Ph.D. from IIT Kharagpur. Kharagpur. His
current research interests include real time computer
vision and its applications.
\end{IEEEbiography}

\begin{IEEEbiography}[{\includegraphics[width=1in,height=1.25in,clip,keepaspectratio]{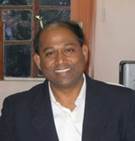}}]{Aurobinda Routray}
is a professor in the
Department of Electrical Engineering, Indian
Institute of Technology, Kharagpur. His research
interest includes non-linear and statistical signal
processing, signal based fault detection and
diagnosis, real time and embedded signal processing,
numerical linear algebra, and data driven
diagnostics.
\end{IEEEbiography}

\vfill

\end{document}